\documentclass[10pt,twocolumn,letterpaper]{article}
\pdfoutput=1 
\usepackage{cvpr}
\usepackage{times}
\usepackage{wrapfig,epsf}
\usepackage{epsfig,epstopdf}
\usepackage{graphicx}
\usepackage{amsmath,amssymb,amsthm}
\usepackage{enumerate}
\usepackage{flushend}
\usepackage{subfigure}
\usepackage[ruled]{algorithm}
\usepackage{algorithmic}

\usepackage{color}
\newtheorem{theorem}{Theorem}
\newtheorem{definition}{Definition} 
\usepackage[pagebackref=true,breaklinks=true,letterpaper=true,colorlinks,bookmarks=false]{hyperref}

\cvprfinalcopy % *** Uncomment this line for the final submission

\def\cvprPaperID{****} % *** Enter the CVPR Paper ID here

\ifcvprfinal\fi
\begin{document}

%%%%%%%%% TITLE
\title{Deep Density-based Image Clustering}

\author{Yazhou Ren, Ni Wang, Mingxia  Li, Zenglin Xu\\
SMILE Lab, School of Computer Science and Engineering, \\University of Electronic Science and Technology of China, \\Chengdu 611731, China\\
{\tt\small yazhou.ren@uestc.edu.cn}
}

\maketitle
%\thispagestyle{empty}

%%%%%%%%% ABSTRACT
\begin{abstract}
%Image clustering is a crucial topic in computer vision and machine learning. 
Recently, deep clustering, which is able to perform feature learning that favors clustering tasks via deep neural networks, has achieved remarkable performance in image clustering applications. However, the existing deep clustering algorithms generally need the number of clusters in advance, which is usually unknown in real-world tasks. In addition, the initial cluster centers in the learned feature space are generated by $k$-means. This only works well on spherical clusters and probably leads to unstable clustering results. In this paper, we propose a two-stage deep density-based image clustering (DDC) framework to address these issues. The first stage is to train a deep convolutional autoencoder (CAE) to extract low-dimensional feature representations from high-dimensional image data, and then apply t-SNE to further reduce the data to a 2-dimensional space favoring density-based clustering algorithms. The second stage is to apply the developed density-based clustering technique on the 2-dimensional embedded data to automatically recognize an appropriate number of clusters with arbitrary shapes. Concretely, a number of local clusters are generated to capture the local structures of clusters, and then are merged via their density relationship to form the final clustering result. Experiments demonstrate that the proposed DDC achieves comparable or even better clustering performance than state-of-the-art deep clustering methods, even though the number of clusters is not given.
\end{abstract}

%%%%%%%%% BODY TEXT
\section{Introduction}\label{sec:introduction}
Image clustering is one of the extensively exploited topics in computer vision and has many applications in a wide range of fields, including image retrieval~\cite{Chen2005Clue, Xie2015IntegratingIC} and annotation~\cite{Li2006Realtime}. 
It seeks to partition images into clusters according to a similarity measure, such that similar images are grouped in the same cluster and images which are dissimilar from each other are grouped into different clusters. 
A number of traditional clustering methods have been proposed in the past decades, such as partitional clustering (e.g., $k$-means \cite{MACQUEEN1967SomeMF}), hierarchical clustering \cite{Jain:clustering}, density-based clustering (e.g., DBSCAN \cite{ester1996density}, mean shift clustering \cite{Dorin2002mean,Ren2014Boosted}), distribution-based clustering (e.g., Gaussian mixture model \cite{Bishop2006Pattern_SR}), etc. 
These methods typically fail to clustering image data sets which are with high dimensionality. The main reason is that reliable similarity measures are hard to obtain in the high dimensional space. 

To mitigate this issue, a normal method is to first reduce the dimensionality of data via feature selection or feature extraction techniques, and then conduct clustering in the lower dimensional space. Another way is to consider clustering and feature learning together in the clustering framework, such as Torre et al. performs $k$-means clustering and linear discriminant analysis jointly \cite{Fernando2006DCA}.
However, these shallow models are typically with limited representation power and thus their improvement on image clustering performance is not significant.

Recently, deep clustering methods, which perform feature learning by applying deep neural networks (DNN) and conduct clustering in the latent learned feature space, have shown impressive performance in image clustering tasks and have attracted people's increasing attentions \cite{chang2017deep,Guo2018DEC-DA,Peng2016Deep,tian2014learning,Xie2016UnsupervisedDE,Yang2017towards,yang2016joint}. 
Despite the huge success, most of the existing deep clustering methods actually apply a partitional clustering, e.g., $k$-means clustering in the latent learned feature space. This brings the following drawbacks:
(1) The number of clusters must be given in advance, which is usually unknown in practical clustering tasks. (2) The partitional clustering techniques can only find spherical clusters and perform worse on irregular clusters or imbalanced data. (3) The $k$-means like clustering methods have randomness, probably leading to unstable clustering results.

Some methods have been proposed to estimate the number of clusters in deep clustering models \cite{lin2018deep,Shah2018DCC,Wang2018DED}. However, these methods  do not consider the local information of clusters, and do not consider that points with different densities should play different roles in density-based clustering technique.  % lack of cluster interpretation, or directly apply existing density-based clustering techniques which do not consider the local structure of clusters and the importance of data points. 
Thus, the performance of these methods is still not satisfied and two questions are normally raising: \textit{(1) How deep clustering methods effectively find appropriate number of clusters with irregular shape when the number of clusters is not known a-prior? (2) Do we really need to refine the deep neural networks  with the initial cluster assignment?}
%\textit{Two questions are raising when clustering image data: (1) Can deep clustering methods effectively find arbitrary shapes of clusters when the number of clusters is not known a-prior? (2) %It is generally claimed in the literature that refining the encoder of autoencoder with the initial cluster centers generated in the learned space will benefit the clustering performance. Nevertheless, d
%Do we really need to refine the autoencoder with the initial cluster centers if good features have already been learned?}

In this paper, we aim to answer these two questions and propose a novel effective deep density-based clustering (DDC) method for images. Specifically, DDC first learns deep feature representation of data via a deep autoencoder. Second, t-SNE \cite{maaten2008tSNE} is adopted to further reduce the learned features to a 2-dimensional space while preserving the pairwise similarity of data instances. Finally, we develop a novel density-based clustering method which considers both the local structures of clusters and importance of instances to generate the final clustering results. %\textbf{The source code of the proposed DDC is available at \url{http://DDC}}

The contributions of this work are stated as below:
\begin{itemize}
    \item We propose a novel effective density-based technique for deep clustering which can automatically find appropriate number of image clusters with arbitrary shapes.
    \item  DDC is with good cluster visualization and interpretability. Its properties are theoretically and empirically analyzed. Its efficiency and robustness to parameter setting are also empirically verified.
    %It is also efficiency because on refinement of neural networks is needed when the cluster assignment is done in the low dimensional space.
    \item Extensive experiments are conducted to show that DDC becomes the new state-of-the-art deep clustering method on various image clusters discovering tasks when the number of clusters is unknown.  
\end{itemize}

\section{Related work}\label{sec:related}
\subsection{Deep clustering}
Due to the good representation ability, deep neural networks (DNN) have gained impressive achievements in various types of machine learning and computer vision applications \cite{Bengio2009Learning,Bengio2013Learning,Hinton2006A}. Most of the DNN methods focus on supervised problems in which the label information is known. In recent several years, people pay increasing attentions to adopting DNN in unsupervised learning tasks and a number of deep clustering methods have been proposed. 

One kind of deep clustering methods divide the clustering procedure into two stages, i.e., feature learning and clustering. They first perform feature learning via DNN and then apply clustering algorithms in the learned space \cite{Chen2015DeepLW,Peng2016Deep,Shao2015DeepLC,tian2014learning}.
The other kind of deep clustering methods incorporate the abovementioned two stages into one framework. Song et al. \cite{song2013auto} refine the autoencoder such that data representations in the learned space are close to their affiliated cluster centers. 
Xie et al. \cite{Xie2016UnsupervisedDE} propose deep embedded clustering (DEC) to jointly learn the cluster assignment and the feature representations. 
Ren et al. \cite{Ren2019SDEC} propose semi-supervised deep embedded clustering to enhance the performance of DEC by using pairwise constraints. 
Yang et al. \cite{yang2016joint} and Chang et al. \cite{chang2017deep} apply convolutional neural networks (CNN) for exploring image clusters.
Guo et al. \cite{guo2017improved} improve DEC with local structure preservation.
Guo et al. \cite{Guo2018DEC-DA} use data augmentation in the DEC framework and achieve state-of-the-art clustering performance on several image data sets.

\subsection{Density-based clustering}
%Density-based clustering generally performs density estimation for each point and then defines clusters as continuous regions of points with high densities.  
The key advantage of density-based clustering is that the number of clusters is not needed and clusters with arbitrary shape can be found.
Over the past decades, many density-based clustering methods have been developed.
DBSCAN \cite{ester1996density} defines a cluster with points from continuous high-density regions and treats those points in low-density regions as outliers or noises. Inspired by this popular algorithm, a lot of density-based clustering methods have been designed, such as OPTICS \cite{ankerst1999optics}, DENCLUE \cite{hinneburg1998efficient}, DESCRY \cite{angiulli2004descry}, and others \cite{du2016density,gu2017parallel,lv2016efficient,mai2015anytime}.
DenPeak (clustering by fast search and find of density peaks
) \cite{rodriguez2014clustering} is another immensely popular density-based clustering method, which assumes that cluster centers locate in regions with higher density and the distances among different centers should be relatively large. Some improvements of DenPeak have also been made \cite{Liu2017adaptive,mehmood2017clustering,XU2016DenPEHC}.
These methods described above are applied in the original feature space. Thus, their performance for grouping images which are with high dimensionality is not satisfied due to the limited representation ability.

In 2018, several deep clustering methods  \cite{lin2018deep,Shah2018DCC,Wang2018DED} which seek to address the issue of estimating the number of clusters have been proposed, i.e., DDC-UF (deep density clustering of unconstrained faces) \cite{lin2018deep}, DCC (deep continuous clustering) \cite{Shah2018DCC}, and DED (deep embedding determination) \cite{Wang2018DED}. 
However, these methods ignore the local structures in each cluster, and do not allow points to play different roles according to their densities. By contrast, the proposed DDC takes into both the local information of clusters and importance of points account and achieves significant improvements on clustering performance.

\iffalse
\textbf{
DCC
DDC-UF 
As a result, the performance of deep clustering methods without knowing the number of clusters should be further improved. To this end, we present a novel deep density-based clustering model for image clustering as below.
Some methods have been proposed to estimate the number of clusters in deep clustering models \cite{lin2018deep,Shah2018DCC,Wang2018DED}. However, these works either lack of cluster interpretation, or directly apply existing density-based clustering techniques which do not consider the local structure of clusters and the importance of data points.
}
\fi 

\section{Deep density-based image clustering}\label{sec:method}
This section presents the proposed deep density-based image clustering (DDC) in detail. Let $\mathcal{X}=\{x_i\in\mathbb{R}^D\}^n_{i=1}$ denote the image data set, where $n$ is number of data points and $D$ is the dimensionality.
DDC aims at grouping $\mathcal{X}$ into an appropriate number of disjoint clusters without any prior knowledge such as the number of clusters and label information.
DDC is a two-stage deep clustering model which contains two main steps, i.e., deep feature learning which nonlinearly transfers the original features to a low dimensional space, and density-based clustering which automatically recognizes an appropriate number of clusters with shapes in the latent space.

\subsection{Deep feature learning}
\label{sec:FeatureLearning}
As deep clustering methods generally do, we adopt deep autoencoder to initialize the feature transformation due to its excellent representation ability.
An autoencoder is consisted of two parts: the encoder $h=f_\Theta(x)$ (maps each data point $x$ to a learned representation $h$) and the decoder $x'=g_\Omega(h)$ (transfers data from the learned feature space to the original one). Here, the feature dimensionality of $h$ is $d$. $\Theta$ and $\Omega$ denote the parameters of the encoder and decoder, respectively.
In this paper, we use the denoising autoencoder \cite{vincent2008} that solves the following problem: %can avoid identity mapping by solving the following problem:
\begin{equation}
\arg\min_{\Theta,\Omega} \frac{1}{n}\sum_{i=1}^{n}\|x_i-g_\Omega(f_\Theta(\tilde{x}_i)) \|_2^2
\label{eq:AE}
\end{equation}
where $\tilde{x}$ is a corrupted copy of $x$ by adding noises, e.g., adding Gaussian noise or randomly setting a portion of input data to $0$. 
We use the stacked autoencoder (SAE) \cite{vincent2010stacked} in this work, in which each layer is a denoising autoencoder trained to reconstruct the previous layer's output.
For image clustering, we adopt the deep convolutional autoencoder (CAE) in the experiments, whose structure will be stated in Section \ref{sec:Implementation}.

%\textbf{Say something about the SAE, stacked autoencoder. We choose deep neural networks (DNN) to initialize the nonlinear transformation $f_\theta$ due to its better representation ability. Concretely, we initialize the DNN structure with SAE, the same as what DEC \cite{Xie2016UnsupervisedDE} does. Each layer of the network is a denoising auto-encoder \cite{vincent2010stacked} trained to reconstruct the previous layer's output after random corruption. After training we concatenate all the encoder and decoder layers together to form a deep auto-encoder \cite{vincent2008}. Please refer to \cite{Xie2016UnsupervisedDE} for more details.}

In \cite{Guo2018DEC-DA}, the data augmentation (DA) technique is used in the training process of deep autoencoder and has achieved significant improvements of clustering performance. The resulting optimization model is:
\begin{equation}
\arg\min_{\Theta,\Omega} \frac{1}{n}\sum_{i=1}^{n}\|\bar{x}_i-g_\Omega(f_\Theta(\bar{x}_i)) \|_2^2
\label{eq:AE-DA}
\end{equation}
where $\bar{x}_i=T_{rand}(x_i)$ denotes the random transformation \footnote{As in \cite{Guo2018DEC-DA}, we randomly shift for at most 3 pixels in each direction and randomly rotate for at most 10$^\circ$.} of $x_i$.

When the training of deep autoencoder (solving Eq. (\ref{eq:AE}) or Eq. (\ref{eq:AE-DA})) is finished, we observe the feature representations $\mathcal{H}=\{h_i=f_\Theta(x_i)\in\mathbb{R}^{d}\}^n_{i=1}$. For visualization and better fitting the designed density-based clustering algorithm, we further reduce data $\mathcal{H}$ to a 2-dimensional space $\mathcal{Z}=\{z_i\in \mathbb{R}^2\}^n_{i=1}$ by using t-SNE \cite{maaten2008tSNE} which owns good preservation ability of pairwise similarities. 
Then, we develop a novel density-based clustering in the embedded space $\mathcal{Z}$ as below. 

\subsection{Density-based clustering}
\label{sec:density-based}
We propose a novel density-based clustering method to obtain an appropriate partition of data $\mathcal{Z}=\{z_i\in \mathbb{R}^2\}^n_{i=1}$ in the 2-dimensional feature space when the number of clusters is unavailable. 

\subsubsection{Local clusters generation}
DDC shares two fundamental definitions (i.e., $\rho_i$ and $\delta_i$ of point $z_i$) with DenPeak \cite{rodriguez2014clustering}. Concretely, DDC defines the density of $\rho_i$ of point $z_i$ via a Gaussian kernel: 
\begin{equation} \label{eq:density}
\rho_i = \sum_{z_j\in\mathcal{Z}\setminus\{z_i\}}\exp\left(-(\frac{d_{ij}}{d_c})^2\right)
\end{equation}
where $d_{ij}$ is the Euclidean distance between points $z_i$ and $z_j$, and $d_c$ is the cutoff distance that need to be predefined.
A higher value of $\rho_i$ means a higher density of point $z_i$.
$\delta_i$ of point $z_i$ denotes the minimum Euclidean distance between $z_i$ and those points whose densities are larger than $z_i$. That is, 
\begin{equation}\label{eq:rho}
\delta_i = \min_{j:\rho_j > \rho_i}(d_{ij})
\end{equation}

For the point with the highest density, its $\rho$ is set to the maximum of pairwise distances.
DenPeak simply chooses several points with the highest $\rho$ and $\delta$ values as cluster centers. Different from DenPeak, we consider those points with relatively large $\rho$ and $\delta$ values as local cluster centers. The corresponding definition is given in Definition \ref{def:local}.
\begin{definition}\label{def:local}
\textbf{(Local cluster centers)} \\Those points satisfying the following condition are defined as local cluster centers:
\begin{equation}\label{eq:localcenter}
\delta_i>d_c \textit{~~and~~} \rho_j>\bar{\rho}
\end{equation}
where $\bar{\rho}=\frac{1}{n}\sum_{j=1}^{n}\rho_j$ is the average density of all the points $\{z_i\}_{i=1}^n$. 
\end{definition}
It is easy to verify that a local cluster center $z_i$ owns the largest density in its $d_c$-neighborhood, i.e., a circle with $z_i$ and $d_c$ as the center and radius, respectively. 
When all the local cluster centers are obtained, we assign each remaining point to the cluster as its nearest neighbor of higher density. Then, a set of local clusters are found and will be used to generate the final clustering. To analyze the characteristic of local cluster centers, the following two theorems are stated.

\begin{theorem}\label{the:center_density}
A local cluster center $z_i$ owns the largest density value $\rho_i$ locally in its $d_c$-neighborhood.
\begin{proof}
We use `proof by contradiction' method to prove the theorem.
For a local cluster center $z_i$, assume that there exists a point $z_j$ in the $d_c$-neighborhood of $z_i$ satisfying $\rho_j>\rho_i$. Then, $\delta_i\le d_c$ holds according to Eq. (\ref{eq:rho}). This actually contradicts Eq. (\ref{eq:localcenter}) in Definition \ref{def:local}. Thus, the assumption is wrong and the theorem is proved.
\end{proof}
\end{theorem}

\begin{theorem}\label{the:center_distance}
The distance of two local cluster centers with different densities is at least $d_c$.
\begin{proof}
Suppose $z_i$ and $z_j$ are two local cluster centers with $\rho_i\ne\rho_j$. We assume the distance $d_{ij}<d_c$, then $z_i$ and $z_j$ are in the $d_c$-neighborhoods of each other.
Since $z_i$ is a local cluster center, it owns the highest density in its $d_c$-neighborhood. Thus, $\rho_i\ge \rho_j$.
$z_j$ is also a local cluster center. Similarly, we have $\rho_j\ge \rho_i$.
Thus, $\rho_i= \rho_j$. This contradicts the condition of the theorem.
\end{proof}
\end{theorem}

Thus, the distance of two local clusters is smaller than $d_c$ only when they have the same density and Eq. (\ref{eq:localcenter}) holds at the same time. In real tasks, this situation extremely rarely occurs.
As a consequence, Theorems \ref{the:center_density} and \ref{the:center_distance} indicate two important properties of local cluster centers: (1) Each local center is with the highest density locally.  (2) The selected cluster centers are not too close to each other, preventing a huge number of cluster centers from being selected.

\subsubsection{Merging local clusters}
Suppose $L$ local clusters ($\mathcal{C}^{(1)},\mathcal{C}^{(2)},\ldots,\mathcal{C}^{(L)}$) are obtained, they will be merged to form the final clustering result.
First, we define core and border points in Definition \ref{def:core}.
\begin{definition}\label{def:core}
\textbf{(Core and border points of a cluster)} \\Suppose a point $z_i$ is from local cluster $\mathcal{C}^{(k)}$, it is defined as a core point if the following condition holds:
\begin{equation}\label{eq:core}
\rho_j>\bar{\rho}^{(k)}
\end{equation}
where $\bar{\rho}^{(k)} = \frac{1}{n_k}\sum_{z_j\in{\mathcal{C}^{(k)}}}\rho_j$ is the average density of all the points in $\mathcal{C}^{(k)}$ and $n_k$ is the number of points in $\mathcal{C}^{(k)}$. Otherwise, $z_i$ is considered as a border point.  
\end{definition}
\noindent Definition \ref{def:core} indicates that whether a point is a core or border point depends on its own density and the average density of the local cluster to which this point belongs.
Generally, the core points of a cluster locate in the central regions, while the border points place in the boundary of areas with lower density.

Then, we define connectivity of clusters in Definitions \ref{def:d-connectivity} and \ref{def:connectivity}.
\begin{definition}\label{def:d-connectivity}
\textbf{(Density directly-connectable of clusters)} \\A local cluster $\mathcal{C}^{(k)}$ is density directly-connectable from a local cluster $\mathcal{C}^{(l)}$ if:
\begin{equation}
\exists~ \text{core points~} z_i\in\mathcal{C}^{(k)} \text{~and~} z_j\in\mathcal{C}^{(l)}, \text{~such that~} d_{ij}< d_c.
\end{equation}
\end{definition}

\begin{definition}\label{def:connectivity}
\textbf{(Density connectable of clusters)} \\A local cluster $\mathcal{C}^{(k)}$ is density-connectable to a local cluster $\mathcal{C}^{(l)}$ if:
\begin{equation}
\exists \text{~a path~} \mathcal{C}^{(k)}=\mathcal{C}_1,\mathcal{C}_2,\ldots, \mathcal{C}_m=\mathcal{C}^{(l)}
\end{equation}
where cluster $\mathcal{C}_{j}$ is density directly-connectable from cluster $\mathcal{C}_{j-1}$ ($j=2,\ldots,m$) and $m$ is the path length.
\end{definition}
\noindent It is easy to verify that both density directly-connectable and density connectable are symmetric. Finally, all the density-connectable local clusters are merged and the final clustering result is provided. 
When two local clusters are merged, the cluster center with higher density becomes the center of the new merged cluster.

According to Definitions \ref{def:d-connectivity} and \ref{def:connectivity}, two clusters are merged only when their central areas are very close to each other. This ensures the new merged cluster also has continuous high-density areas. % and reduces the risk of the incorrect merging when noises data and 

\iffalse
\begin{theorem}\label{the:center_density33}
If two clusters are density directly-connectable of clusters and thus be merged. The density of overlapped areas is not lower than $\min\{p_i,p_j\}$. 
\textbf{This ensures that continuous high-density areas of the merged cluster. 
This prevents merging two clusters with overlapped areas which are with low density.
A local cluster center $z_i$ owns the largest density value $\rho_i$ locally in its neighborhood.}
\begin{proof}
The
\end{proof}
\end{theorem}
\fi

The pseudo-code of the proposed DDC is summarized in Algorithm \ref{Alg:DDC}.

%\textbf{The flowchart of the proposed DDC is summarized in Fig. 1}.

\begin{algorithm}[!t]
\caption{Deep Density-based Image Clustering (DDC).}
\label{Alg:DDC}
\begin{algorithmic}[1]
\REQUIRE ~~%~\\input
Image data set $\mathcal{X}$; Cutoff distance $d_c$.
\ENSURE %~\\output
The final clustering result. 
\STATE \textbf{Stage 1} $\rightarrow$ \textbf{Deep feature learning}
\STATE Train a deep autoencoder via Eq. (\ref{eq:AE}) or (\ref{eq:AE-DA}).
\STATE Transform $\mathcal{X}$ to lower feature representations $\mathcal{H}$ via the encoder $f_\Theta(\cdot)$.
\STATE Map $\mathcal{H}$ to a 2-dimensional data set $\mathcal{Z}$ via t-SNE.
\STATE \textbf{Stage 2} $\rightarrow$ \textbf{Density-based clustering}
%\STATE \% \textbf{Local clusters generation}
\FOR {each point $z_i$ in $\mathcal{Z}$}
\STATE Compute $\rho_i$ and $\delta_i$ via Eqs. (\ref{eq:density}) and (\ref{eq:rho}).
\ENDFOR
\STATE Choose local cluster centers via Eq. (\ref{eq:localcenter}).
\STATE Assign the remaining points and observe local clusters $\mathcal{C}^{(1)},\mathcal{C}^{(2)},\ldots,\mathcal{C}^{(L)}$.
%\STATE \% \textbf{Generate the final clustering result}
\STATE Define core and border points via Eq. (\ref{eq:core}).
\STATE Merge all the density connectable local clusters.
\STATE Return the final clustering result.
\end{algorithmic}
\end{algorithm}

\begin{figure*}[!t]
\centering
\subfigure[Decision graph]{\includegraphics[width=0.4\columnwidth]{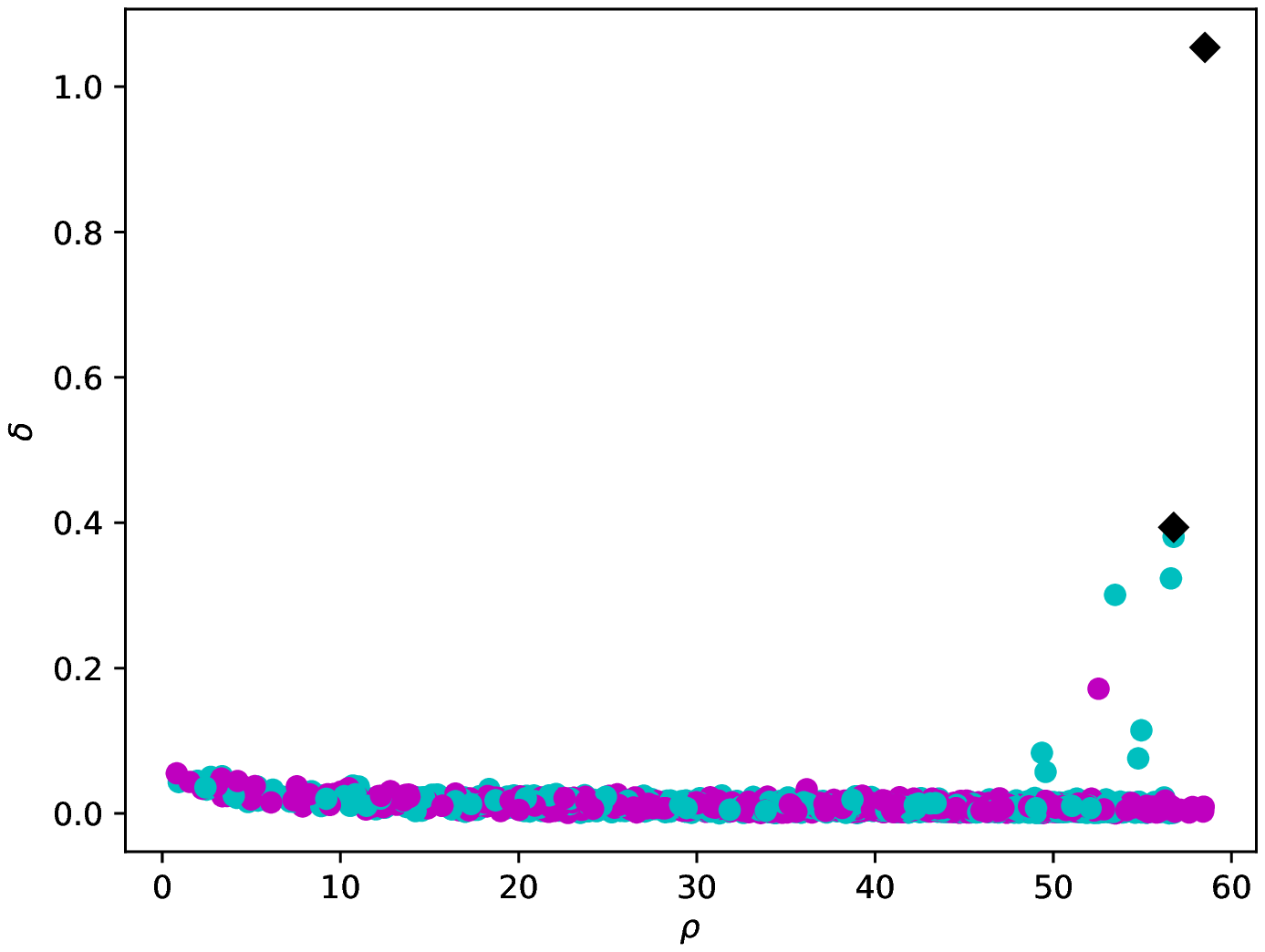}}
\subfigure[Clustering result]{\includegraphics[width=0.4\columnwidth]{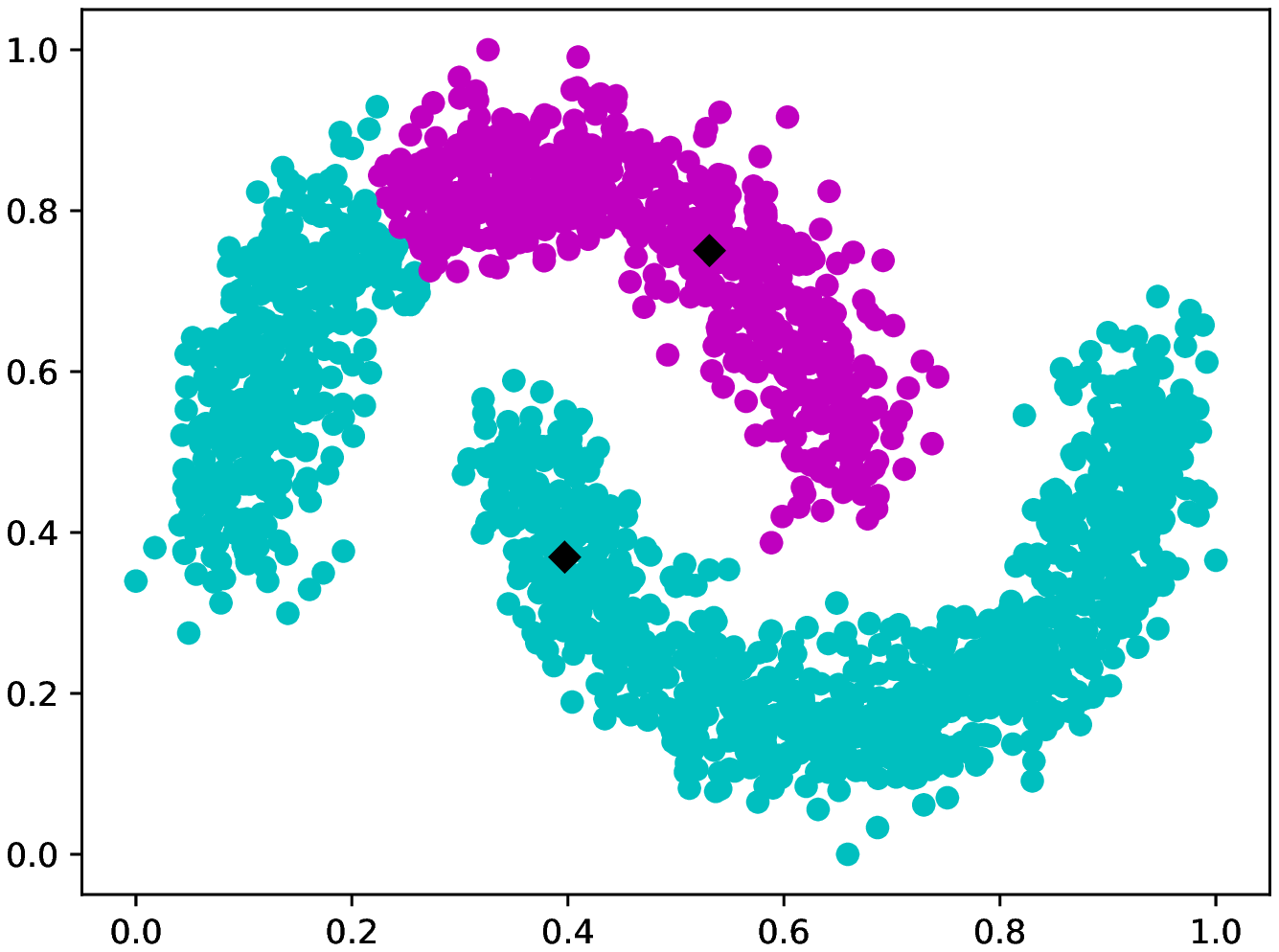}}
\subfigure[Initial result]{\includegraphics[width=0.4\columnwidth]{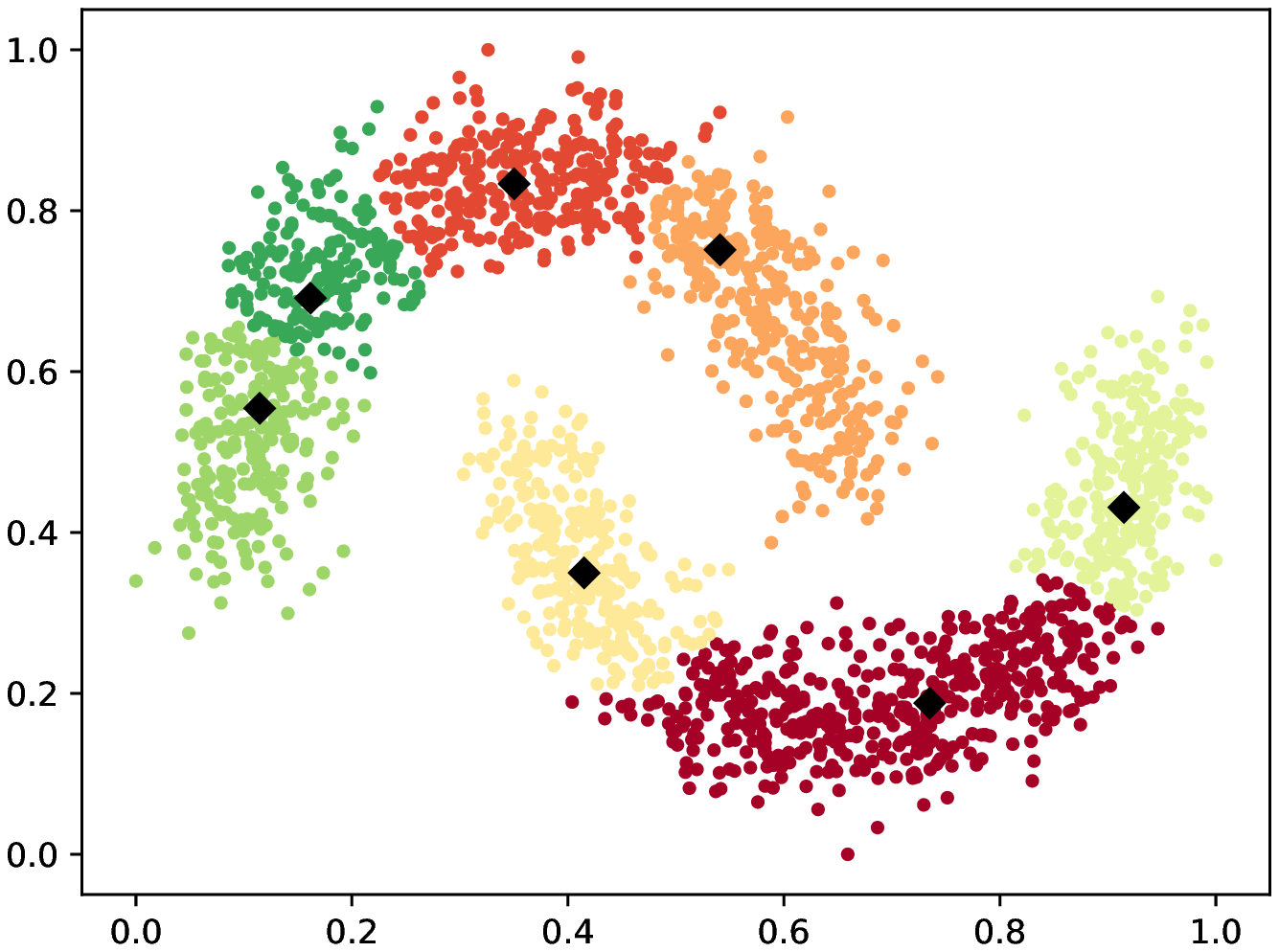}}
\subfigure[Final result]{\includegraphics[width=0.4\columnwidth]{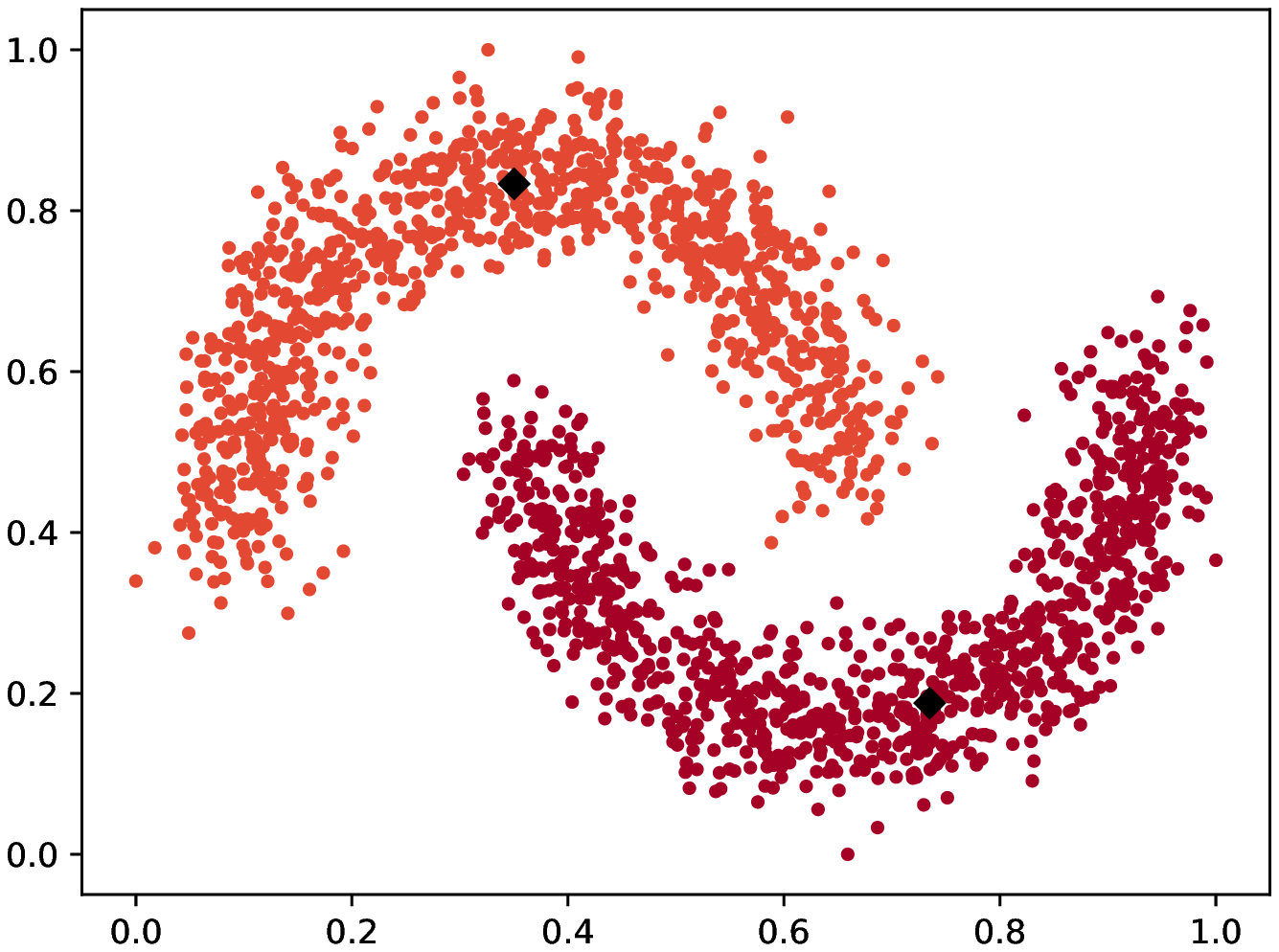}}
\subfigure[Border points]{\includegraphics[width=0.4\columnwidth]{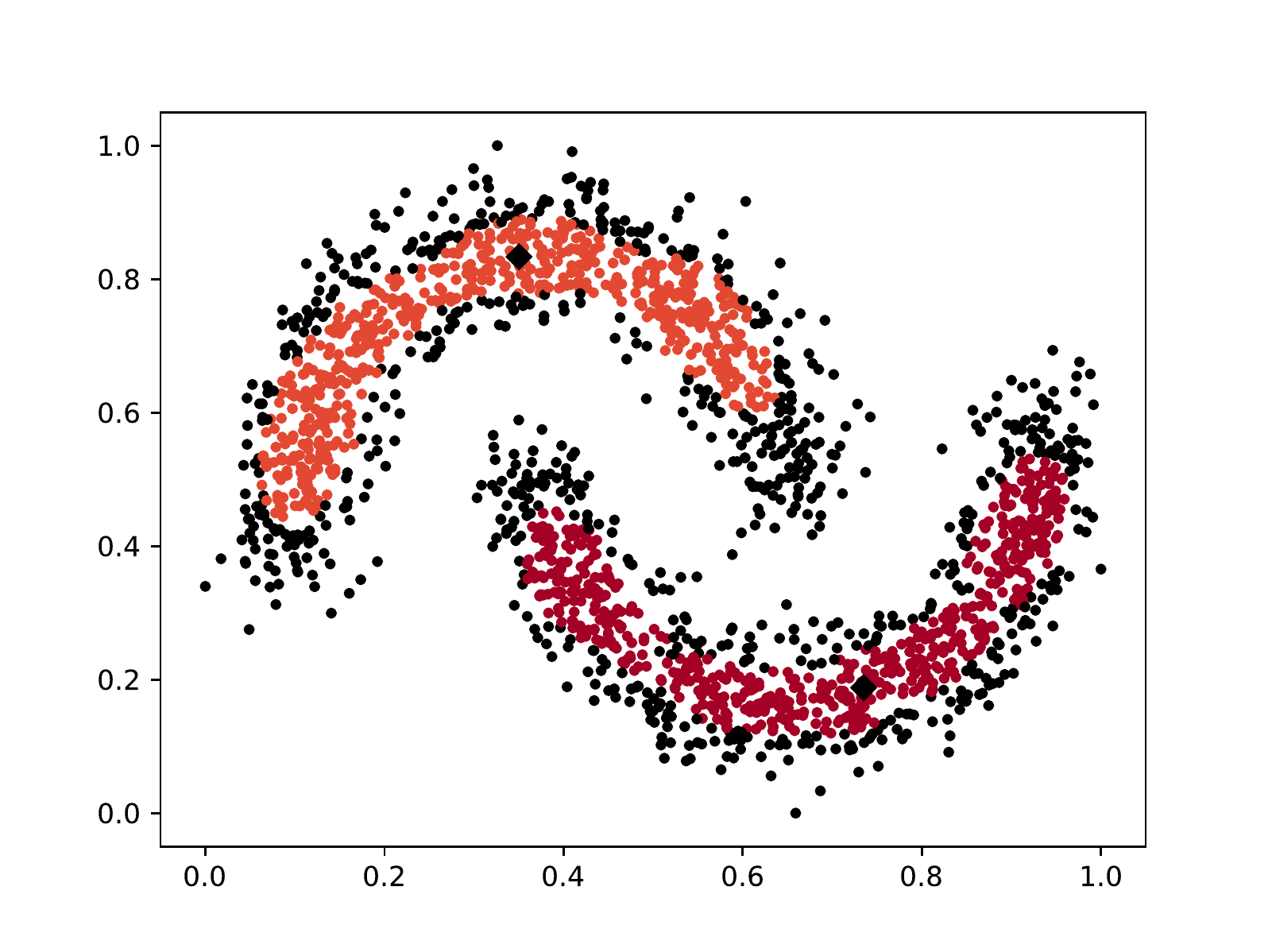}}
\caption{Twomoon: Clustering performance comparison of DenPeak and DDC. The Twomoon data set has 2000 points from two classes. (a): The decision graph of DenPeak. (b): The final result of DenPeak. (c): Initial local clusters of DDC. (d): The final result of DDC. (e): The border points detected by DDC are plotted as black points. The center of each cluster is highlighted with black `$\scriptscriptstyle\blacklozenge$'. 
Points with the same color are from the same cluster.
As shown in (a), a number of points with high $\rho$ and $\delta$ values can be considered as centers and it is hard for DenPeak to choose an appropriate number of clusters. Even it is told that 2 clusters exist, the result of DenPeak is still not satisfied, as (b) shows. By contrast, DDC first generate a relatively large number of local cluster centers and then merge them to form the final clustering result. Compared (c) with (e), we find that 
two clusters are typically merged if there exists core points that are from both clusters and are close to each other.
It is shown in (e) that border points generally locate around the boundary of each real cluster, while core points locate in central areas.
}
\label{fig:res_DenPeak_DDC}
\end{figure*}

\setcounter{footnote}{0}

\begin{figure*}[!t]
\centering
\subfigure[Result of DenPeak]{\includegraphics[width=0.5\columnwidth]{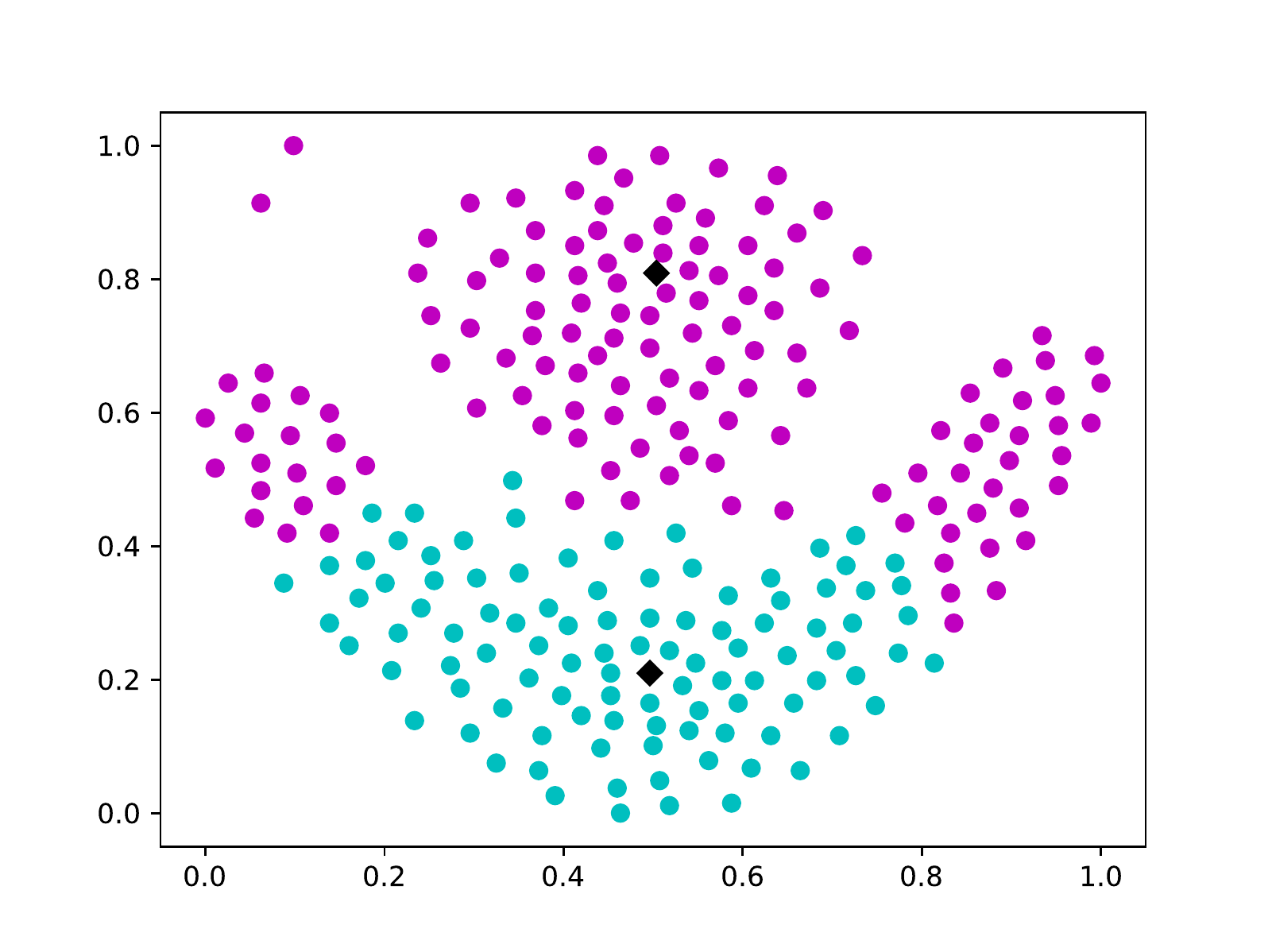}}
\subfigure[Result of DDC]{\includegraphics[width=0.5\columnwidth]{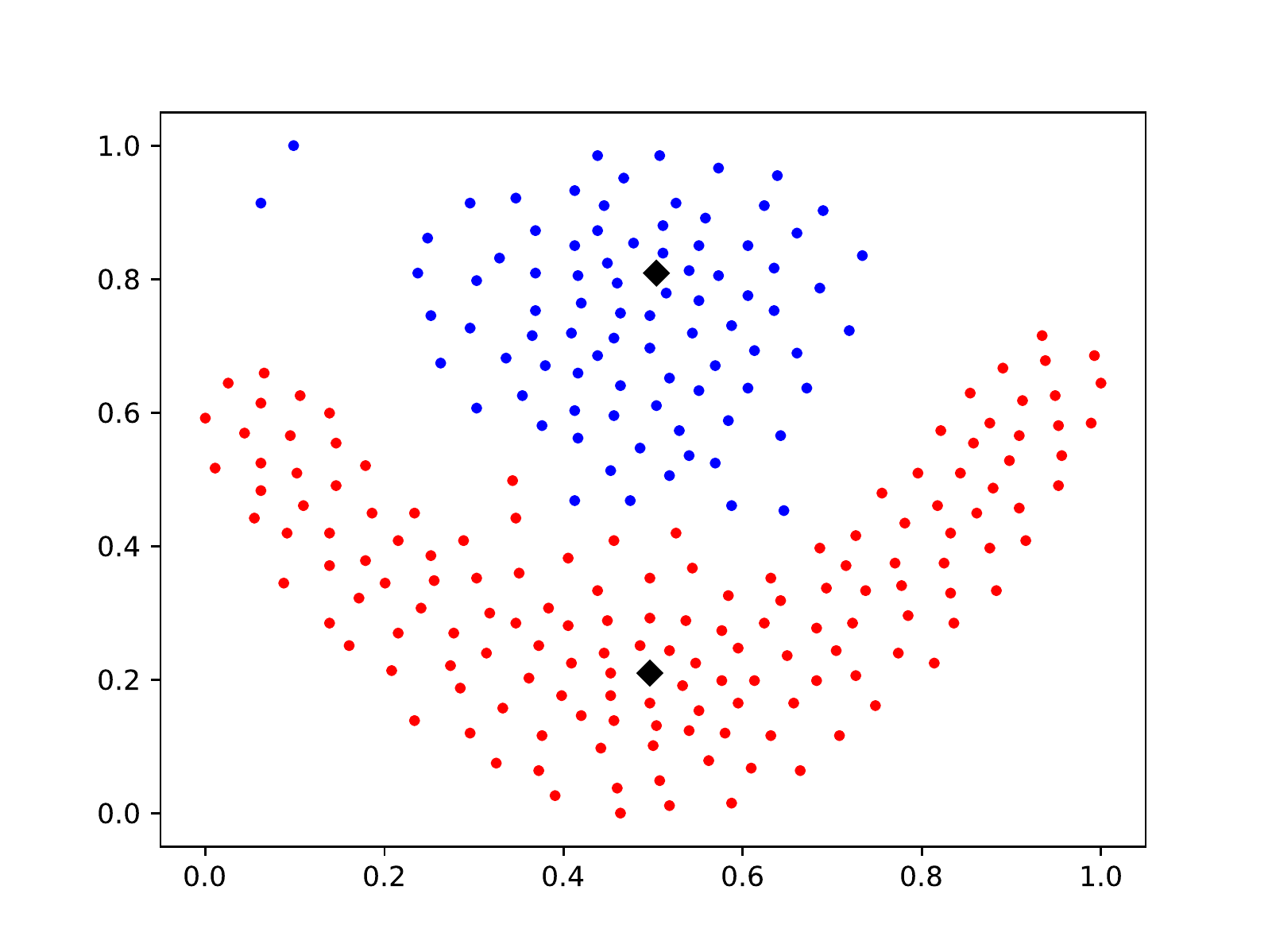}}
\subfigure[Result of DenPeak]{\includegraphics[width=0.5\columnwidth]{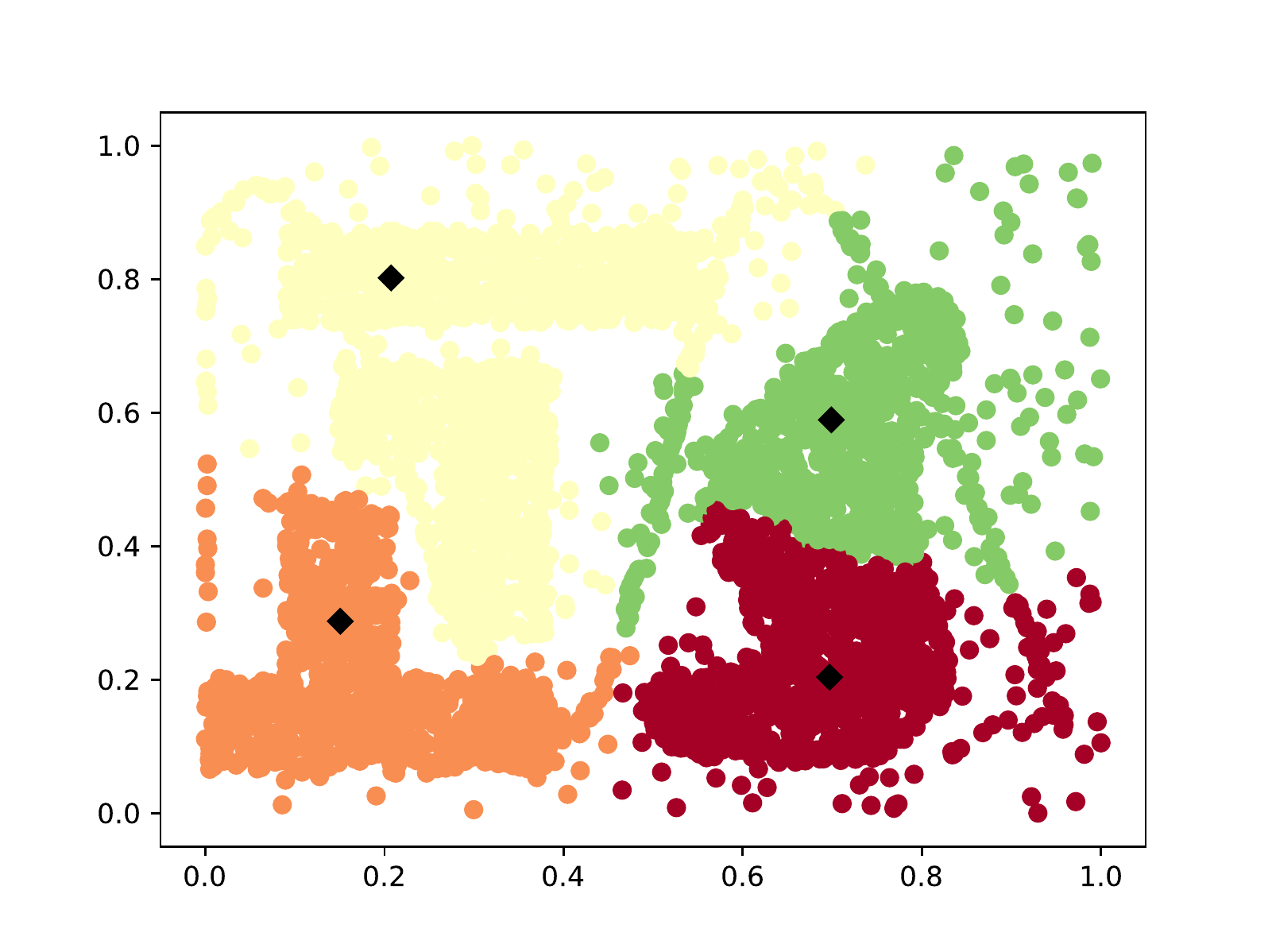}}
\subfigure[Result of DDC]{\includegraphics[width=0.5\columnwidth]{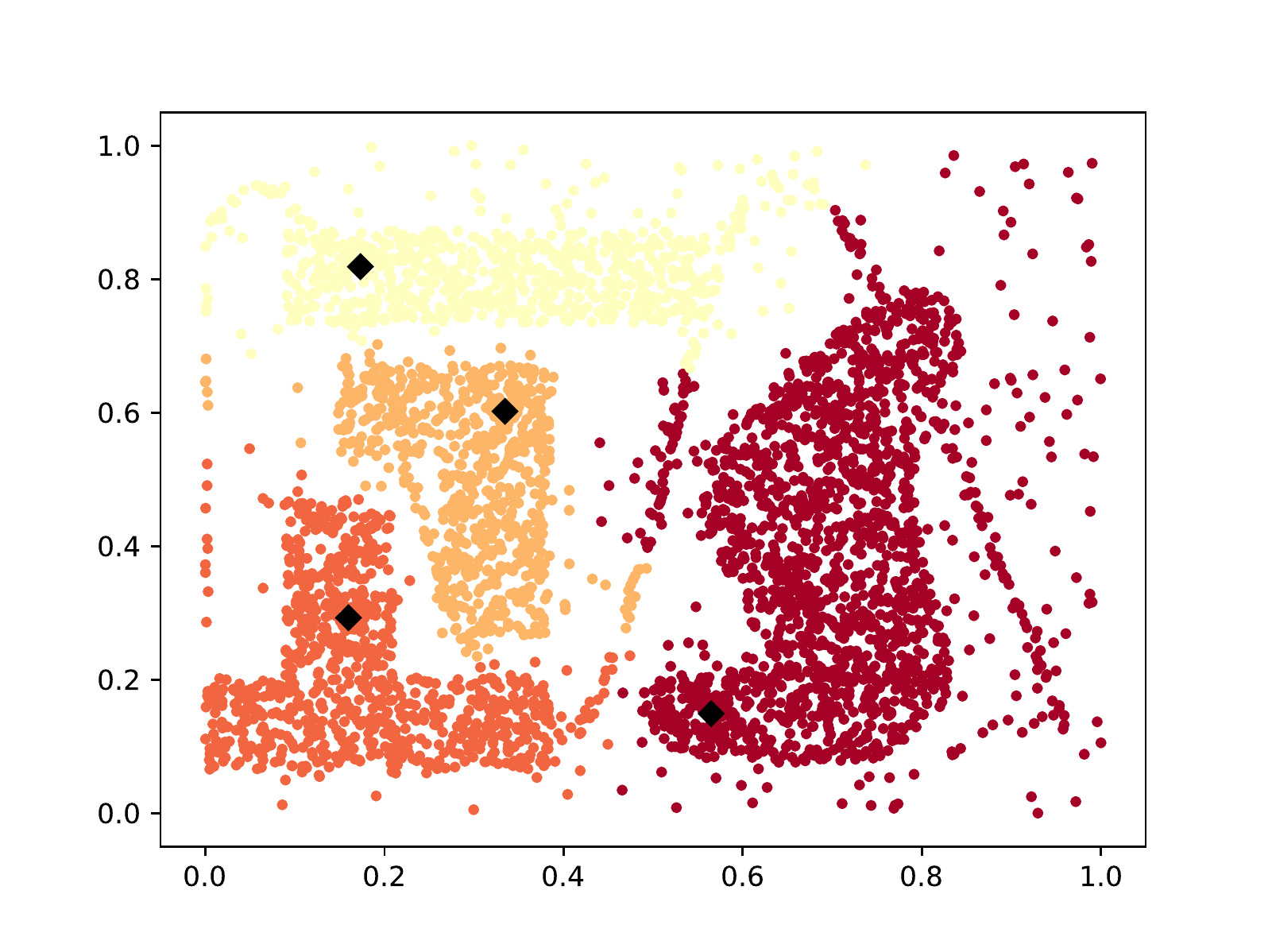}}
\caption{Clustering results of DenPeak and DDC on Flame and t4 data sets \protect\footnotemark. (a) and (b) correspond to the Flame data set. (c) and (d) show the results on t4. DenPeak is told to select the true number of clusters. Due to loss information of local structures, DenPeak fails to find suitable clusters (as shown in (a) and (c)). In contrast, DDC performs perfectly on these two data sets. Even when noisy data exist (as exhibited in (d)), DDC can still automatically recognize the 4 irregular clusters.}
\label{fig:res_DenPeak_DDC2}
\end{figure*}

\subsection{Implementation}\label{sec:Implementation}
According to different optimization problems, DDC provides two specific algorithms:
\begin{enumerate}[(1)] %\setlength{\itemsep}{0pt}
%\item Fc-DDC: Use full connected deep autoencoder and solving Eq. (\ref{eq:AE}).
%\item Fc-DDC-DA: Use full connected deep autoencoder and solving Eq. (\ref{eq:AE-DA}) where the data augmentation (DA) is adopted.
\item DDC: Use CAE and solve Eq. (\ref{eq:AE}).
\item DDC-DA: Use CAE and solve Eq. (\ref{eq:AE-DA}) in which data augmentation is adopted.
\end{enumerate}

%The structure of a full connected autoencoder is always set to $D-500-500-2000-10-2000-500-500-D$, where $D$ is the feature dimensionality of original data. 
The structure of encoder layers of CAE is always set to $\text{Conv}_{32}^{5}\rightarrow\text{Conv}_{64}^{5}\rightarrow\text{Conv}_{128}^{3}\rightarrow\text{Fc}_{10}\rightarrow\text{Conv}_{128}^{3}\rightarrow\text{Conv}_{64}^{5}\rightarrow\text{Conv}_{32}^{5}$.  Here, $\text{Conv}_{32}^{5}$ represents a convolutional layer with $32$ filters and a $5\times 5$ kernel.  The stride is always set to 2. $\text{Fc}_{10}$ denotes the full connected layer with $10$ neurons.
In convolutional autoencoders, all the internal layers except for the input, embedding, and output layers are activated by ReLU function.
The structures of autoencoders also indicate that the dimensionality of learned representations $\mathcal{H}$ is 10.

%\subsection{Parameter selection}
Given the embedded 2-dimensional data $\mathcal{Z}$, DDC has only one parameter ($d_c$) needed to be set. 
We set the value of $d_c$ according to data $\mathcal{Z}$ itself. Concretely, we compute $\bar{d}$ as the average value of all pairwise distances in $\mathcal{Z}$. Then, set $d_c=\bar{d}\times ratio$. 
If $ratio$ is extremely large, a small number of clusters will be found by DDC. If $ratio$ is extremely small, a large number of clusters will be detected. 
However, we will empirically verify that DDC achieves stable performance in a wide range of $ratio$. The default value of $ratio$ is $0.1$.

\footnotetext{\url{http://cs.joensuu.fi/sipu/datasets/}}

\subsection{Relations to exiting methods}
DBSCAN \cite{ester1996density} and DenPeak \cite{rodriguez2014clustering} are two worldwide popular density-based clustering methods. They are applied in the original feature space, while the proposed DDC works in the 2-dimensional embedded space. Besides, DBSCAN is sensitive to the parameters and tends to merge clusters with overlapping areas \cite{ester1996density,Ren2014Boosted}. These shortcomings prevent its successful use in image clustering tasks.

DenPeak assumes that each cluster has only one center, leading to the following disadvantages: (1) In real applications, multiple centers/modes usually coexist in one cluster. Thus, DenPeak typically loses information of local structures of a cluster. (2) It is difficult for DenPeak to select a suitable number of clusters because usually a number of (which is much larger than the ground-truth number of clusters) points with high $\rho$ and $\delta$ values can be considered as candidates of cluster centers. 
To address these issues, DDC firstly selects all the potential cluster centers to obtain the local clusters, and then aggregates all density connectable cluster to form the final clustering result. An illustration exhibiting the different behaviors of DenPeak and DDC is given in Figs. (\ref{fig:res_DenPeak_DDC}) and (\ref{fig:res_DenPeak_DDC2}). 
Here, DCC is directly applied on the 2-dimensional data without using CAE and t-SNE. %The $ratio$ of DDC is 0.1. 
DenPeak follows the parameter setting described in Section \ref{sec:methods}.
%More empirical study may be found in supplementary materials.

DED \cite{Wang2018DED} is a recently proposed deep clustering model that transforms the original data via DNN to a 2-dimensional feature space that favors the density-based clustering algorithm. However, DED directly applies DenPeak on the 2-dimensional data, thereby inheriting the disadvantages of DenPeak.

\section{Experimental setup}\label{sec:exp_set}
This section describes the tested image data sets, comparing methods, parameter settings, and evaluation measures.

\subsection{Image data sets}
\begin{table}[!h]
\caption{Image data sets used in the experiments.}
\centering%\begin{center}
\begin{tabular}{cccc}
\hline
data set & \# examples & \# classes & image size\\
\hline
MNIST & 70000 & 10 & 28$\times$28 \\
MNIST-test & 10000 & 10 & 28$\times$28 \\
USPS & 9298 &10 & 16$\times$16  \\
Fashion & 10000  & 10  & 28$\times$28 \\
LetterA-J  & 10000  & 10  & 28$\times$28 \\
%STL-10 & 13000 &10 &1428\\
%CIFAR-10 & 60000 & 10 & 180\\
%20NG  &  2965 & 3  &7270\\
\hline
\end{tabular}
%\end{center}
\label{tab:data}
\end{table}
Five popular image data sets are used to assess the performance of comparing methods.

The MNIST data base \footnote{\url{http://yann.lecun.com/exdb/mnist/}} consists of 70000 handwritten digits of 28$\times$28 pixel size from 10 categories (digits 0-9). 
The MNIST-test data set only contains the test set of MNIST, with 10000 images.
The USPS data set \footnote{\url{https://www.csie.ntu.edu.tw/~cjlin/libsvmtools/datasets/multiclass.html}} is collected from handwritten digits from envelopes by the U.S. postal service. It contains 9298 grayscale images with size $16\times 16$. 
Fashion \cite{Xiao2017Fashion} is a data set comprising $28\times 28$ gray images of 70000 fashion products from 10 categories. Its test set with 10000 images are used in our experiments.
The LetterA-J data set \footnote{\url{https://yaroslavvb.blogspot.com/2011/09/notmnist-dataset.html}} is consisted of more than 500k $28
\times 28$ greyscale images of English letters from A to J. We randomly select 10000 images from its uncleaned subset as test set.

The summary of all data sets is shown in Table \ref{tab:data}. The features of each data set are scaled to $[0,1]$. 
%When applying fully connected autoencoder, each image is reshaped as a $D$-dimensional vector.

\subsection{Evaluation measures}\label{sec:metric}
Clustering accuracy (ACC) and normalized mutual information (NMI) are used to estimate the performance of comparing algorithms. Their values are both in [0,1]. A higher value of ACC or NMI indicates a better clustering performance.

\subsection{Comparing methods}\label{sec:methods}
We compare the proposed DDC with both shallow clustering methods and deep ones.
Shallow baselines are $k$-means \cite{MACQUEEN1967SomeMF}, DBSCAN \cite{ester1996density}, and DenPeak \cite{rodriguez2014clustering}. 
Deep methods based on both full connected and convolutional autoencoders are compared, including DEC (deep embedded clustering) \cite{Xie2016UnsupervisedDE}, IDEC (improved DEC with local structure preservation) \cite{guo2017improved}, DCN (deep clustering network) \cite{Yang2017towards}, JULE (joint unsupervised learning for image clustering) \cite{yang2016joint}, DCC (deep continuous clustering) \cite{Shah2018DCC}, DED (deep embedding determination) \cite{Wang2018DED}, DEC-DA (DEC with data augmentation) \cite{Guo2018DEC-DA}.

Among all the comparing methods, DBSCAN, DenPeak, DCC, DED, and the proposed DCC do not need the number of clusters in advance. For all other methods, the number of clusters is set to the the ground-truth number of categories. 
When applying DBSCAN, the 4-th nearest neighbor distances are computed w.r.t. the entire data, and parameter $Eps$ is set to the median of those values. The $MinPts$ value of DBSCAN is always set to 4. For DenPeak, the Gaussian kernel is used and $d_c$ is set such that the average number of points in $d_c$-neighborhood is approximately $1\% \times n$. To give DenPeak and DED an advantage, the detected number of clusters is set to the true number of classes according to the decision graph.
%DED shares the same convolutional structure with our DDC, and DEC-DA uses the same structures of both fully connected and convolutional autoencoders with DDC.
So far, given the ground-truth number of clusters, ConvDEC-DA achieves state-of-the-art clustering performance in image clustering \cite{Guo2018DEC-DA}. We compare ConvDEC-DA and its version without using DA in our experiments.

The reported ACC and NMI values are either excerpted from the original papers, or are the average values of running the released code with corresponding suggested parameters for 10 independent trials.

\section{Results and analysis}\label{sec:results}

\subsection{Results on real image data}

\begin{table*}[!t]
  \centering
  \caption{Results of the comparing methods. In each column, the best two results are highlighted in boldface. The results marked by `*' are excerpted from the papers. `-' denotes the results are unavailable from the papers or codes, and `- -' means `out of memory' when applying.}
 % that the corresponding running time is not tolerated ($>$24 hours).}
    \begin{tabular}{c|cccccccccc}
    \hline
          & \multicolumn{2}{c}{MNIST} & \multicolumn{2}{c}{MNIST-test} & \multicolumn{2}{c}{USPS} & \multicolumn{2}{c}{Fashion} & \multicolumn{2}{c}{LetterA-J} \\
          & ACC   & NMI   & ACC   & NMI   & ACC   & NMI   & ACC   & NMI   & ACC   & NMI \\
    \hline
    $k$-means & 0.485  & 0.470  & 0.563  & 0.510  & 0.611  & 0.607  & 0.554  & 0.512  & 0.354  & 0.309  \\
    DBSCAN & - -     & - -     & 0.114  & 0     & 0.167  & 0     & 0.1   & 0     & 0.1   & 0 \\
    DenPeak & - -     & - -     & 0.357  & 0.399  & 0.390  & 0.433  & 0.344  & 0.398  & 0.300  & 0.211  \\
    DEC   & 0.849  & 0.816  & 0.856  & 0.830  & 0.758  & 0.769  & 0.591  & 0.618  & 0.407  & 0.374  \\
    IDEC  & 0.881* & 0.867* & 0.846  & 0.802  & 0.759  & 0.777  & 0.523  & 0.600  & {0.381 } & {0.318 } \\
    DCN   & 0.830* & 0.810* & 0.802* & 0.786* & 0.688* & 0.683* & -     & -     & -     & - \\
    JULE  & 0.964* & 0.913* & 0.961* & 0.915* & 0.950* & 0.913* & -     & -     & -     & - \\
    DCC   & 0.963*  & -     & -     & -     & -     & -     & -     & -     & -     & - \\
    DED   & - -     & - -     & 0.690  & 0.818  & 0.781  & 0.855  & 0.473  & 0.617  & 0.371  & 0.440  \\
    ConvDEC & 0.940  & 0.916  & 0.861  & 0.847  & 0.784  & 0.820  & 0.514  & 0.588  & 0.517  & 0.536  \\
    ConvDEC-DA & \textbf{0.985 } & \textbf{0.961 } & {0.955 } & {\textbf{0.949 }} & {\textbf{0.970 }} & {\textbf{0.953 }} & 0.570  & 0.632  & 0.571  & \textbf{0.608}  \\
    DDC & {0.965 } &{0.932 } & \textbf{0.965 } & 0.916  & 0.967  & 0.918  & \textbf{0.619 } & \textbf{0.682 } & \textbf{0.573 } & {0.546 } \\
    DDC-DA & {\textbf{0.969 }} & {\textbf{0.941 }} & \textbf{0.970 } & \textbf{0.927 } & \textbf{0.977 } & \textbf{0.939 } & \textbf{0.609 } & \textbf{0.661 } & \textbf{0.691 } & \textbf{0.629 } \\
    \hline
    \end{tabular}%
  \label{tab:results}%
\end{table*}%

%The default value of parameter $ratio$ (in Section \ref{sec:Implementation}) of DDC is set to 0.1.  
Table \ref{tab:results} gives the clustering results of comparing methods measured by ACC and NMI. In each column, the best two results are highlighted in boldface. 
From Table \ref{tab:results} we have the following observations: (1) The shallow models generally perform worse than deep clustering methods. DBSCAN works the worst mainly because it is hard to choose suitable parameters in high dimensional space.
(2) Data augmentation (DA) can improve the clustering performance. Except for two methods using DA (i.e., ConvDEC-DA and DDC-DA), our DDC always achieves the highest ACC and NMI values.
(3) Our DDC-DA always achieves one of the best two clustering results, even the number of clusters is not given.
Even given the true number of clusters, DED still performs much worse than DDC and DDC-DA.
(4) We also find that ConvDEC-DA can usually obtain a high ACC value ($>$0.98), but it performs worse (ACC $<$0.84) occasionally. This might be caused by the bad initial cluster centers provided by $k$-means in the learned feature space. By contrast, our DDC and DDC-DA are more stable with small standard deviations. % (which are not reported due to limited space). 

\begin{table}[!t]
  \centering
  \caption{The average number of detected clusters.}
    \begin{tabular}{ccc}
    \hline
      data set    & DDC   & DDC-DA \\
    \hline
    MNIST & 10.8$\pm$0.4 & 10.7$\pm$0.5 \\
    MNIST-test & 10.0$\pm$0.0 & 10.0$\pm$0.0 \\
    USPS  & 10.0$\pm$0.0 & 10.0$\pm$0.0 \\
    Fashion & 10.5$\pm$0.8 & 10.2$\pm$0.8 \\
    LetterA-J & 9.1$\pm$0.8 & 10.1$\pm$0.9 \\
    \hline
    \end{tabular}%
  \label{tab:numberK}%
\end{table}%
The average number of clusters detected by our DDC and DDC-DA as well as the corresponding standard deviations are given in Table \ref{tab:numberK}. From Table \ref{tab:numberK} we find that our methods can always find the correct numbers of categories on MNIST-test, and USPS. On MNIST, Fashion and LetterA-J, the recognized numbers of clusters are slightly different from the true values. These indicate the capability of the proposed DDC framework of automatically recognizing reasonable numbers of clusters.
%On MNIST, a small set of points sometimes gather together and form an independent cluster. Thus, the resulting detected number of clusters is 11. However, this does not affect the clustering performance heavily, as shown in Table \ref{tab:results}.

\subsection{Sensitivity analysis}\label{sec:sensitivity}
\begin{figure}[!t]
\centering
\subfigure[MNIST-test]{\includegraphics[width=0.495\columnwidth]{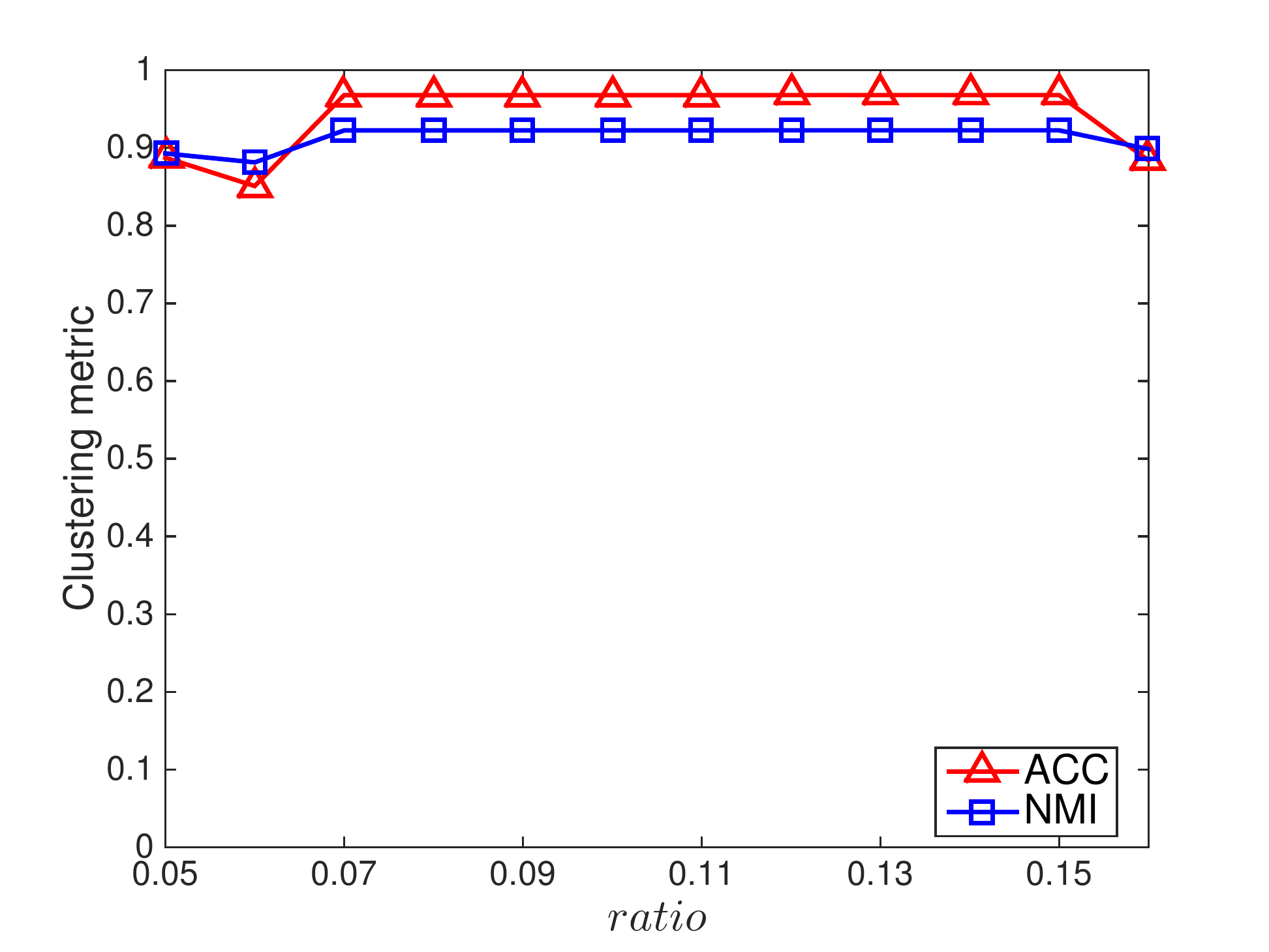}}
\subfigure[USPS]{\includegraphics[width=0.495\columnwidth]{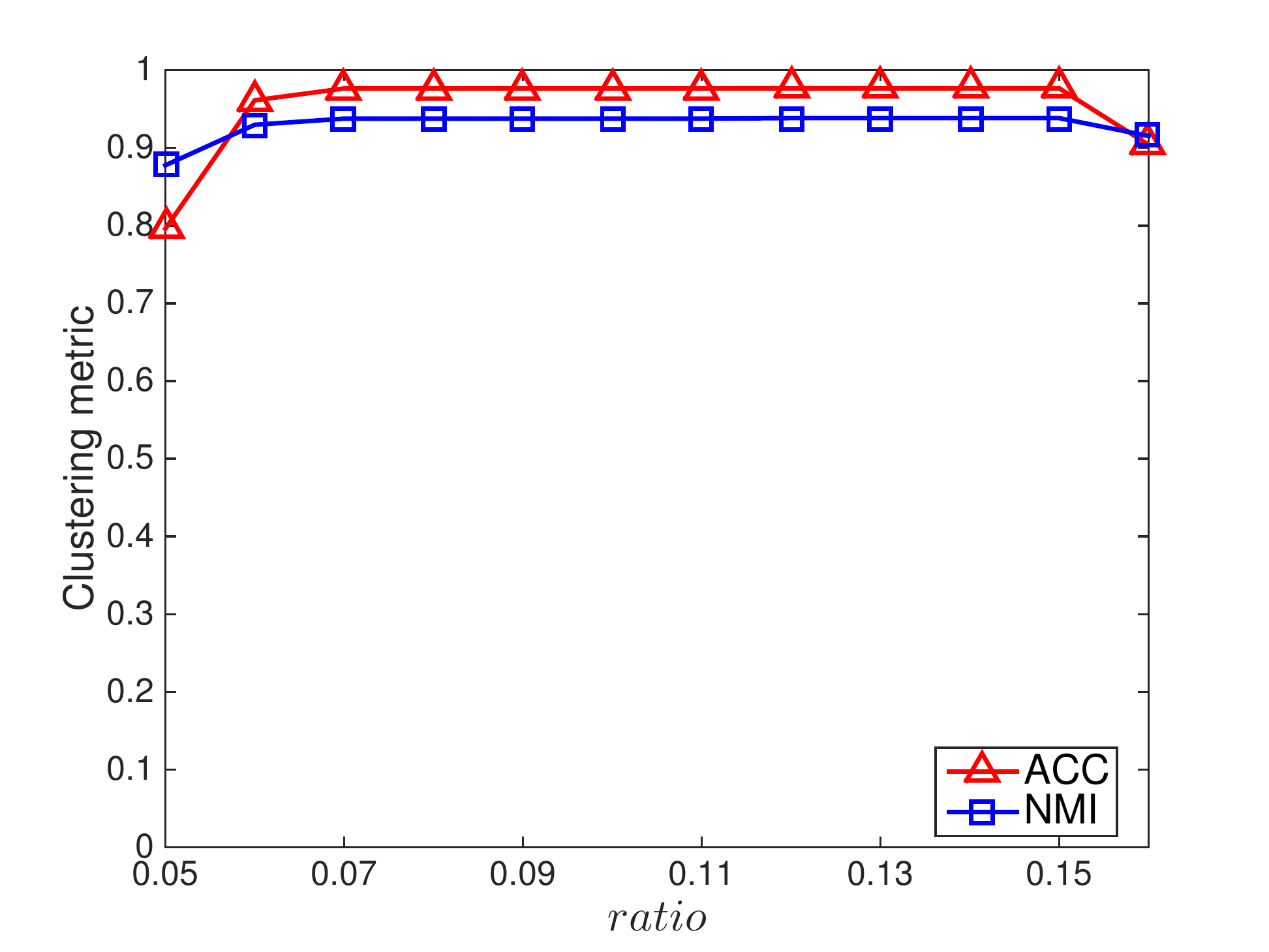}}
\caption{Sensitivity analysis of parameter $ratio$ (ACC and NMI).}
\label{fig:sensitivity}
\end{figure}

This section tests the sensitivity of DCC w.r.t. the parameter $ratio$ on MNIST-test and USPS data sets. 
The tested range is $[0.05,0.16]$. Both ACC and NMI values of DDC-DA are reported in Fig. \ref{fig:sensitivity}, from which we can observe that our method achieves stably excellent performance in a wide range of $ratio$. When applying the DDC methods in real clustering applications, the default value of $ratio$ is recommended to be set to 0.1.

\subsection{Runtime analysis}
We compare our method with DEC-DA \cite{Guo2018DEC-DA} because these two models use the same CAE structure and DEC-DA has been proved to be efficient compared with other existing deep clustering methods. The experiments are tested on a server with 32 GB RAM and 2 Tesla P100 GPUs. 
Concretely, the runtimes of our DDC-DA on MNIST-test and USPS are 737 and 583 seconds, respectively. Those of ConvDEC-DA are 798 and 436 seconds, respectively. DDC-DA needs time to estimate the density $\rho$ and $\delta$ for each point. ConvDEC-DA needs to refine the CAE with initial cluster centers. Thus, these two methods show competitive performance in terms of efficiency. 
%However, our method does not need to know the number of clusters. Furthermore, the distance and density calculations of DDC can be further reduced with $k$-means 

\begin{figure*}[!t]
\centering
\subfigure[Ground truth labels]{\includegraphics[width=0.495\columnwidth]{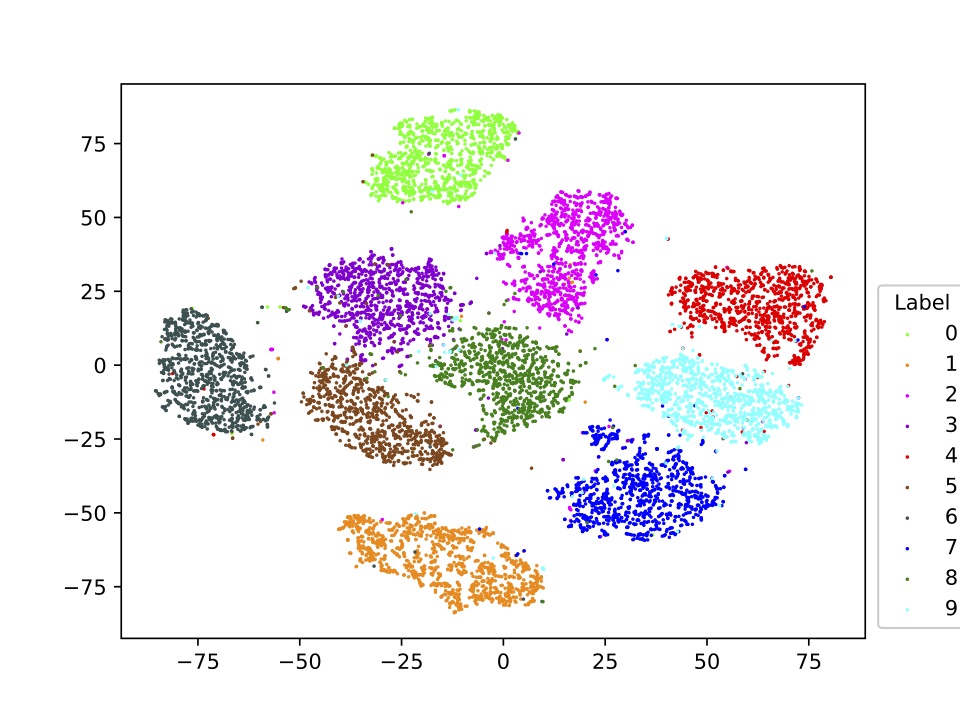}}
\subfigure[Initial result]{\includegraphics[width=0.495\columnwidth]{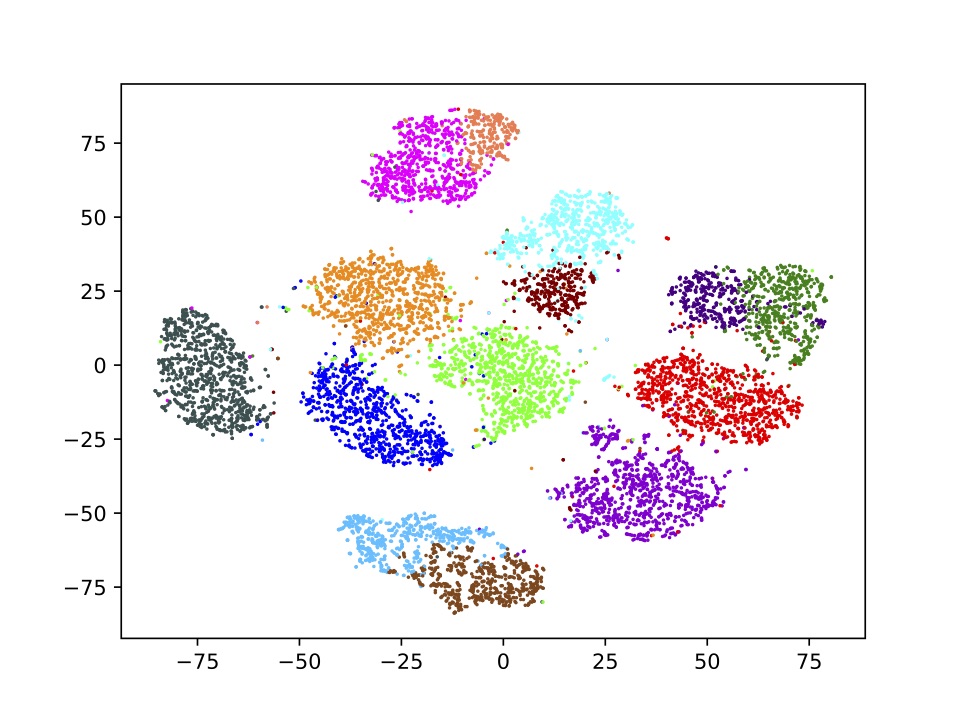}}
\subfigure[Final result]{\includegraphics[width=0.495\columnwidth]{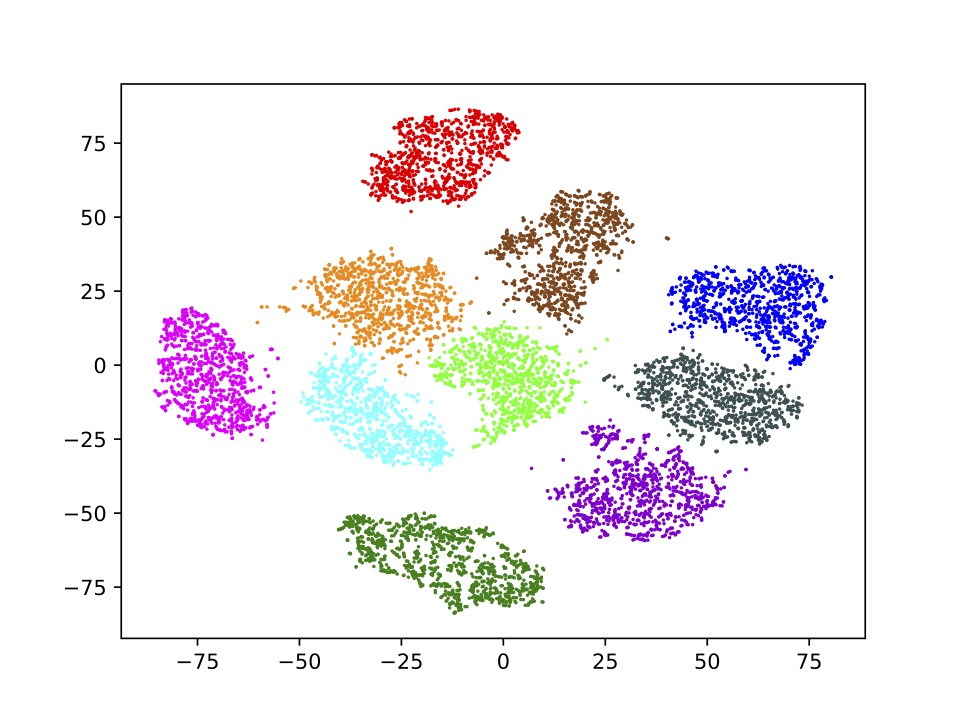}}
\subfigure[Border points]{\includegraphics[width=0.495\columnwidth]{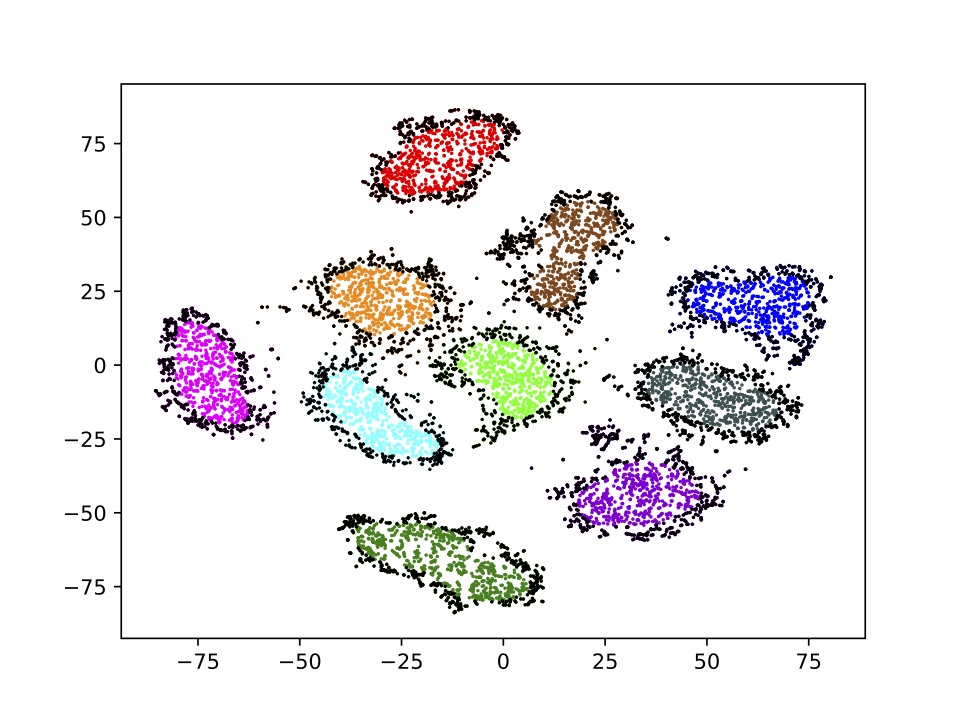}}
\caption{Visualization of DDC-DA on MNIST-test. (a) The ground truth labels of the embedded 2-dimensional data. (b) The initial result of DDC-DA. (c) The final result of DDC-DA. (d) The border points detected by DDC-DA.}
\label{fig:MNIST_test}
\end{figure*}

\begin{figure*}[!t]
\centering
\subfigure[Ground truth labels]{\includegraphics[width=0.495\columnwidth]{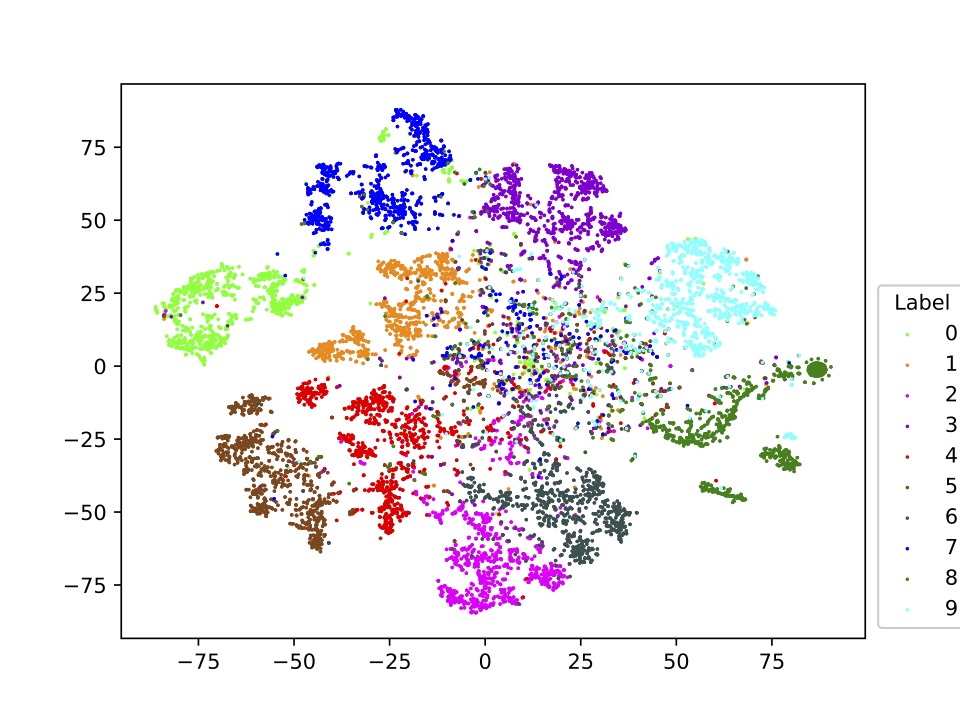}}
\subfigure[Initial result]{\includegraphics[width=0.495\columnwidth]{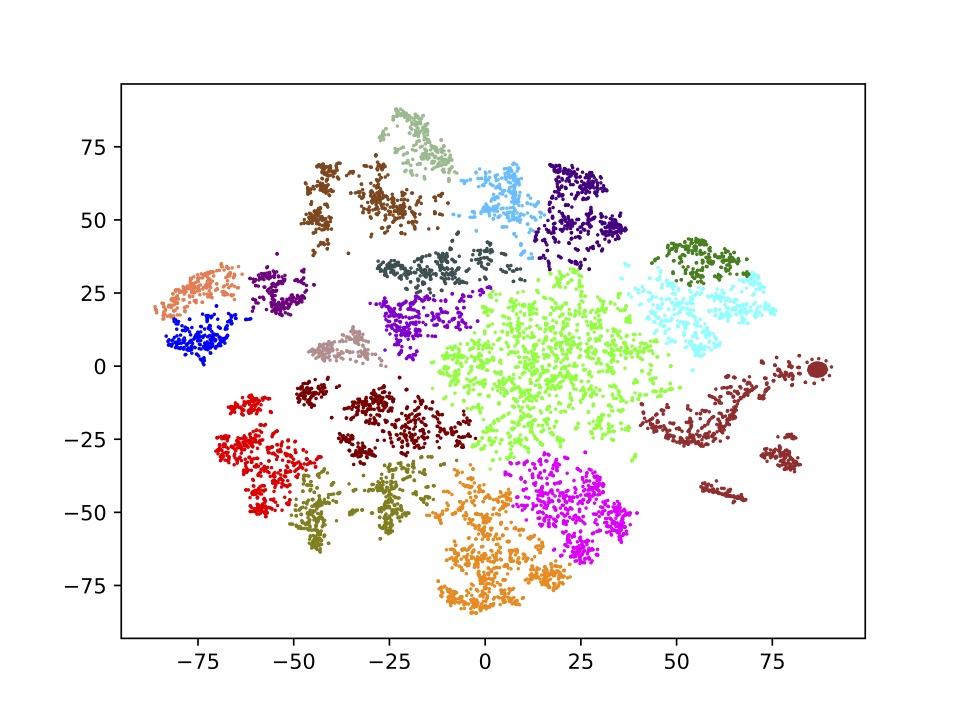}}
\subfigure[Final result]{\includegraphics[width=0.495\columnwidth]{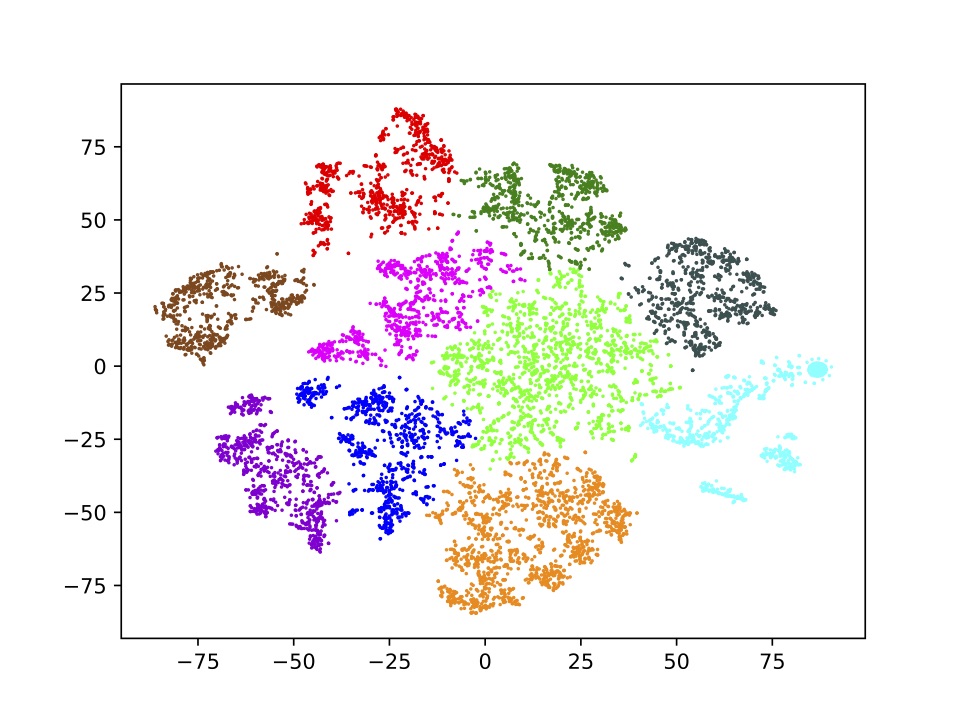}}
\subfigure[Border points]{\includegraphics[width=0.495\columnwidth]{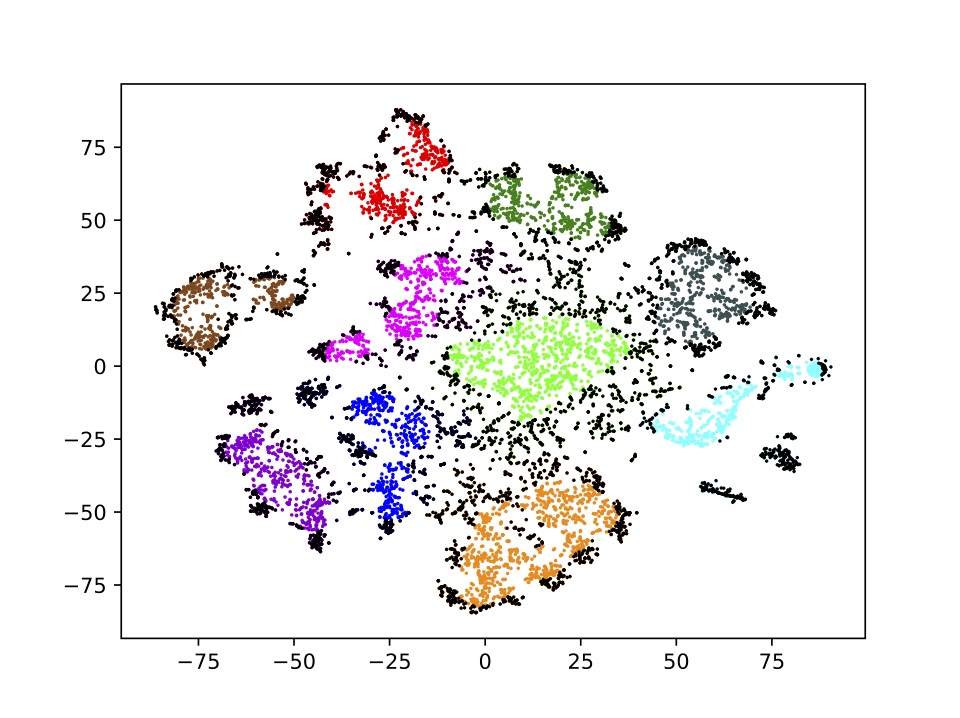}}
\caption{Visualization of DDC-DA on LetterA-J. (a) The ground truth labels of the embedded 2-dimensional data. (b) The initial result of DDC-DA. (c) The final result of DDC-DA. (d) The border points detected by DDC-DA.}
\label{fig:nonmnist}
\end{figure*}

\section{Discussion}\label{sec:Discussion}
We also conduct experiments to directly use t-SNE to reduce the original data to the 2-dimensional space and then apply the proposed density-based clustering technique. The clustering results are much worse than our DDC methods. The main reason is that CAE can transform the original data to a lower dimensional space in which the intrinsic local structures are preserved. It is better to further reduce the lower dimensional representations to a 2-dimensional space rather than extracting from the original high dimensional data. 
%which preserve intrinsic local structure better models the distribution of the true categories. 
As a consequence, DED \cite{Wang2018DED} and our DDC make use of both CAE and t-SNE to obtain the 2-dimensional representations that favor the density-based clustering.

Now, let us come back to the question raised in Section \ref{sec:introduction}: Is it really needed to refine the deep autoencoder with the initial cluster assignment? 
To answer this question, we first visualize the clustering results on MNIST-test and LetterA-J in the embedded 2-dimensional space of DDC-DA in Figs. \ref{fig:MNIST_test} and \ref{fig:nonmnist}, respectively. 
For data whose clusters are well separated (as shown in Fig. \ref{fig:MNIST_test} (a)), those centroid-based clustering methods, such as ConvDEC-DA, which depends greatly on the initial selection of cluster centers, needs to refine the CAE iteratively to achieve satisfied results.
By contrast, our DDC can output remarkable performance without refinement even when several clusters in the middle area have overlapped areas.

For data in which many points from different categories mess together (as shown in the middle area of Fig. \ref{fig:nonmnist} (a)), the refinement of ConvDEC-DA can not separate the messed points correctly, neither does our DDC. 
If this happens and no additional information is given, the effectiveness of refining autoencoder is not significant for both centroid-based and density-based clustering.
In our opinion, one needs prior information (e.g., pairwise constraints) or knowledge transferred from related tasks to handle this situation.

\section{Conclusion and future work}\label{sec:conclusion}
This article has introduced a novel deep density-based clustering (DDC) method for images. It is well known that for high-dimensional data such as images, it is difficult to obtain satisfied performance by applying clustering methods in the original space of image data. So in DDC, first, we use CAE with good representation ability to extract 10-dimensional features from the original data. After this, t-SNE is used to reduce the 10-dimensional data to a 2-dimensional space, which favors our density-based clustering. DDC consider both the local information of clusters and the importance of points in the clustering process. It is empirically proved to be the new state-of-the-art deep clustering method when the number of clusters is not given. Its efficiency and robustness are also verified. 
An interesting future work is to exploit semi-supervised learning and transfer learning into deep density-based clustering.

%\section*{Acknowledgments}
%This paper was in part supported by Grants from the Natural Science Foundation of China (Nos. 61806043, 61572111, and 61832001), two Projects funded by China Postdoctoral Science Foundation (Nos. 2016M602674 and 2017M623007), a 985 Project of UESTC (No. A1098531023601041), and three Fundamental Research Funds for the Central Universities of China (Nos. ZYGX2016J078, ZYGX2016Z003, and ZYGX2016J034).

{\small
\bibliographystyle{ieee}
\bibliography{Yazhou}
}

\end{document}